\documentclass[10pt, a4paper]{article}

\usepackage{lrec-coling2024}

\pdfoutput=1
\hypersetup{
    breaklinks=true,
    colorlinks=true
}

\usepackage{inconsolata}
\usepackage{amsmath}
\usepackage{amsfonts}
\usepackage{amssymb}
\usepackage{algorithm}
\usepackage[noend]{algpseudocode}
\usepackage{booktabs}
\usepackage{csquotes}
\usepackage{listings}
\usepackage{tipa}

\DeclareFontSubstitution{T3}{cmss}{m}{n}

\usepackage{float}
\usepackage{multirow}
\usepackage{verbatim}
\usepackage[super]{nth}
\usepackage{tikz}
\usepackage{forest}
\usetikzlibrary{fit,backgrounds,calc}
\usepackage[capitalize,nameinlink]{cleveref}

\usepackage{mathastext}

\usepackage{tipa}

\newcommand{\eg}{\textit{e.g.}}
\newcommand{\ie}{\textit{i.e.}}

\newcommand{\nop}[1]{}
\newcommand{\code}[1]{\texttt{#1}}

\newcommand{\magnify}[1]{\textbf{#1}}

\newcommand{\halfquad}{\,\,\,}

\newcommand{\stress}[1]{\'{#1}}

\definecolor{burntorange}{RGB}{179,50,14}
\definecolor{royalpurple}{RGB}{133,64,222}
\definecolor{darkgreen}{RGB}{51,117,11}

\newcommand{\latin}[1]{\textcolor{royalpurple}{\texttt{#1}}}
\newcommand{\pital}[1]{\textcolor{burntorange}{\texttt{#1}}}

\newcommand{\tricolon}{\textipa{:}}
\newcommand{\latinemph}[1]{\magnify{#1}}
\newcommand{\pitalemph}[1]{\magnify{#1}}
\newcommand{\gloss}[1]{`#1'}

\usepackage{mathtools}

\renewcommand{\pm}{\text{±}}
\let\emptyset\varnothing
\renewcommand{\neq}{\mathrel{\mathclap{\mkern7mu\not}=}}

\newcommand{\bos}{\textsc{bos}}
\newcommand{\eos}{\textsc{eos}}
\newcommand{\dmodel}{d_{\mathrm{model}}}

\defcitealias{lrcIELEX2024}{LRC, 2024}

\title{PILA: A Historical-Linguistic Dataset of Proto-Italic and Latin}

\name{Stephen Bothwell,$^\ast$ Brian DuSell,$^\dag$ David Chiang,$^\ast$ Brian Krostenko$^\ast$} 

\address{
\begin{tabular}{cc}
$^\ast$University of Notre Dame & $^\dag$ETH Z\"{u}rich \\
Notre Dame, Indiana, USA & Z\"{u}rich, Switzerland \\
\texttt{\{sbothwel,dchiang,bkrosten\}@nd.edu} & \texttt{brian.dusell@inf.ethz.ch}
\end{tabular}}

\abstract{
 Computational historical linguistics seeks to systematically understand processes of sound change, including during periods at which little to no formal recording of language is attested. At the same time, few computational resources exist which deeply explore phonological and morphological connections between proto-languages and their descendants. This is particularly true for the family of Italic languages. To assist historical linguists in the study of Italic sound change, we introduce the Proto-Italic to Latin (PILA) dataset, which consists of roughly 3,000 pairs of forms from Proto-Italic and Latin. We provide a detailed description of how our dataset was created and organized. Then, we exhibit PILA's value in two ways. First, we present baseline results for PILA on a pair of traditional computational historical linguistics tasks. Second, we demonstrate PILA's capability for enhancing other historical-linguistic datasets through a dataset compatibility study. \\ \newline 
}

\begin{document}

\maketitleabstract

\section{Introduction}
\label{sec:intro}

All languages change over time, but much work in computational linguistics views language as a static phenomenon. Most natural language processing models (\eg, for named entity recognition or machine translation) are trained on snapshots of languages at a fixed point in time. In contrast, historical linguistics---and, by extension, computational historical linguistics---attempts to track linguistic shifts across various points in time. 

The phonological side of historical linguistics examines the relations among \textit{cognate sets}: forms in different languages that are related etymologically. If forms have a common ancestor, or \textit{etymon} (pl. \textit{etyma}), they are known as \textit{reflexes} of that etymon. Historical linguists hypothesize systems of sound change to explain the evolution of language over time. However, as we proceed further into the past, reconstruction becomes more difficult or even speculative due to the dearth of existing evidence.

Consider the examples presented in \cref{tab:sound-change-sample} for the Proto-Italic and Latin languages. Each row corresponds to a sound change pattern. \emph{Cluster reduction}, for instance, involves the collapse of a longer consonant cluster into a shorter one. Here, the reconstructed \pital{*takslos} reduces \pital{*ksl} into \latin{l}.\footnote{In this paper as well as in our dataset, we use phonetic transcriptions for both languages. We detail our transcription scheme in \cref{sec:pila-normalization}.} How does this collapse occur? Many sequences of rules could be proposed to explain the phenomenon. To reconstruct the true process, historical linguists gather evidence by comparing languages in close temporal and geographic proximity. Yet, to our knowledge, such efforts have not demonstrated the precise sequence of rules that underlies cluster reduction---alongside other sound change patterns.

\begin{table}
    \centering
    \small
    \begin{tabular}{@{}llll@{}}
        \toprule
        Pattern & Proto-Italic & Latin & Gloss \\
        \midrule 
        Vowel weakening & \pital{*frag\pitalemph{e}lis} & \latin{frag\latinemph{i}lis} &  \gloss{brittle} \\
        Cluster reduction & \pital{*ta\pitalemph{ksl}os} & \latin{ta\tricolon \latinemph{l}us} & \gloss{ankle} \\
        Syncope & \pital{*a\tricolon z\pitalemph{i}de\tricolon jo\tricolon} & \latin{a\tricolon rdeo\tricolon} & \gloss{I burn} \\
        Metathesis & \pital{*pa\pitalemph{ur}os} & \latin{pa\latinemph{rv}us} & \gloss{few} \\
        \bottomrule
    \end{tabular}
    \caption{Sampling of sound change patterns present in PILA. Phones affected by the listed pattern are underlined and written in boldface. Here and elsewhere we write Proto-Italic forms in orange (with asterisks), and Latin forms in purple.}
    \label{tab:sound-change-sample}
\end{table}

To scrutinize systems of sound change and build models to capture them, we need sizable datasets. Toward that end, we introduce PILA, the \textbf{P}roto-\textbf{I}talic to \textbf{La}tin dataset.\footnote{Our dataset is available at the following location: \url{https://github.com/Mythologos/PILA}.} PILA contributes to the study of historical linguistics in many ways. Namely: 
\begin{itemize}
    \item PILA is the first dataset to contain full (and not partial; see \Cref{sec:pila-trimming}) reconstructions of Proto-Italic etyma and their Latin reflexes.
    \item PILA is one of the largest available datasets of etymon--reflex pairs for a single proto-language. (See \Cref{tab:proto-dataset-comparison}.) 
    \item PILA provides multiple inflections for most lemmata, letting phonological studies consider morphology's influence. (See \Cref{sec:augmentation-inflections}.)
    \item PILA highlights the presence of non-phonological changes (\eg, analogy) through per-entry annotations. (See \Cref{tab:irregularity-table}.)
\end{itemize}

After situating our work in \cref{sec:background,sec:related-work}, we describe our dataset (\cref{sec:pila-overview}) and its development (\cref{sec:pila-development}). Then, in \cref{sec:applications}, we exhibit PILA's applicability through strong baseline results on standard historical linguistics tasks and a careful dataset compatibility study. We further provide a table in our dataset to encourage studies which span multiple historical linguistics datasets. (See \Cref{sec:dataset-compatibility} for details.)

\section{Background}
\label{sec:background}

\begin{figure}
    \centering
    \small
    \begin{forest}
        for tree={parent anchor=south,child anchor=north,s sep=0pt,l=1cm},
        [Indo-European [Italic,name=italic,edge={dashed} [Latino-Faliscan,name=latfal [Latin,name=lat [ Romance,name=romance [Spanish,edge=dashed] 
        [French,edge=dashed]  [Italian,edge=dashed] 
        [{\makebox[3em]{$\cdots$}},edge=dashed] 
        ] ] [Faliscan] ] [Osco-Umbrian] ] ] ]
        \node[fill=burntorange,circle,minimum size=1mm,inner sep=1.5pt,label={[align=center,label distance=0.25cm,name=pilabel,color=burntorange]160:Pre-Latin \\ Proto-Italic}](protoitalic) at ($(latfal.north)!0.35!(italic.south)$) {};
        \draw[burntorange] (pilabel)--(protoitalic);
        \node[fill=royalpurple,circle,minimum size=1mm,inner sep=1.5pt,label={[align=center,label distance=0.3cm,name=ciclabel,color=royalpurple]180:Ciceronian \\ Latin}](ciclatin) at ($(lat.north)!0.05!(latfal.south)$) {};
        \draw [royalpurple] (ciclabel)--(ciclatin);
        \draw[thick,arrows = {->}] (-4.5,.5) -- (-4.5,-5) node[anchor=center,midway,above,rotate=-90] {time};
    \end{forest}
    \caption{Partial family tree of Italic and Latin. Dashed lines indicate the omission of some intermediate families and branching structures from the tree. The two points represent the time periods covered by our dataset.}
    \label{fig:partial-family-tree}
\end{figure}
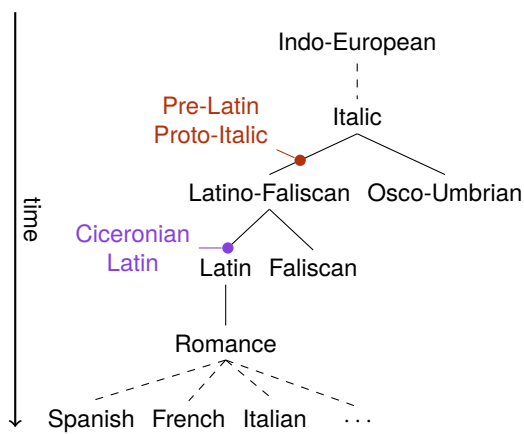

The Proto-Indo-European (abbreviated PIE) language is a reconstructed, unattested language. Historical linguists theorize that as the language spread, it split up into branches characterized by shared innovations. One set of innovations characterizes a branch conventionally named Proto-Italo-Celtic, which is the parent of later Italic and Celtic languages. 
One branch of that family, characterized by further innovations \citep[8]{devaanEtymologicalDictionaryLatin2008}, is dubbed Proto-Italic. That language is the parent of a family of attested languages spoken chiefly in the Italian peninsula, which fall into dialect or family groups, generally along geographical lines. 

The language family centered on the Apennine spine is Osco-Umbrian, divided into Oscan in the south and Umbrian in the north. Meanwhile, the language family originally spoken in the lower country along the Tyrrhenian coast between Etruria and the Campania is Latino-Faliscan, divided into Faliscan in the north and Latin in the south. As the city-state of Rome developed into an international power, Latin spread widely, eventually supplanting the other Italic languages. Thus, Latin is undoubtedly the best attested of the older Italic languages, and it becomes the source of the Romance languages in turn. We depict a summarized version of this language family tree in \cref{fig:partial-family-tree}.

Turning to these languages' phonetic inventories, Proto-Italic inherits a fairly small set of phones from PIE \citep[124]{beekesComparativeIndoEuropeanLinguistics1995}. By the time Proto-Italic splits up, laryngeals---a class of sounds with a guttural articulation---disappear. They leave behind traces of various kinds (\eg, long vowels, the creation of \texttt{a}). Meanwhile, PIE's voiced aspirates generally become fricatives, and their vocalized nasals become nasal stops (\texttt{em}, \texttt{en}). Within Latin the fricatives are heavily reordered, sometimes becoming stops; diphthongs have become or are becoming monophthongs; and the labiovelars sometimes have lost labial or velar articulation. 

PIE is a highly inflected language; Proto-Italic and Latin follow suit, although these languages gradually simplify PIE's system. Like PIE, Proto-Italic and Latin inflect through patterns of suffixation. PIE has a nominal system with stem classes that fall into three agreement categories, conventionally called genders, and can be suffixed into eight or nine cases. By the time of Latin, only five or six cases (\eg, genitive, accusative) are in full use. As for PIE's verbal system, it employs vowel changes in the stem and suffixation variously to assume different aspects, moods (\eg, indicative, imperative), and persons (\nth{1}, \nth{2}, and \nth{3}). Latin largely drops the meaning behind vowel changes in verb stems and instead standardizes these changes into four conjugations. Every verb is characterized by number (singular, plural), person, tense (\eg, present, perfect), mood, and voice (\eg, active, passive).

\section{Related Work}
\label{sec:related-work}

\begin{table*}[t]
    \small
    \centering
    \begin{tabular}{@{}ll@{}}
        \toprule
        Family & Coverage \\
        \midrule
        Arawakan & Purus$^\dag$ \citelanguageresource{decarvalhoComparativeReconstructionProtoPurus2021} \\
        \multirow[t]{4}{7em}{Austronesian} & Austronesian Basic Vocabulary Database$^\dag$ \citelanguageresource{greenhillAustronesianBasicVocabulary2008}, \\
        & \textsc{Jambu} [Munda] \citelanguageresource{aroraJambuHistoricalLinguistic2023}, \\ 
        & Micronesian Comparative Dictionary$^\dag$ \citelanguageresource{benderProtoMicronesianReconstructions2003a,benderProtoMicronesianReconstructions2003b}, \\
        & Tukanoan \citelanguageresource{chaconRevisedProposalProtoTukanoan2014} \\
        Dravidian & \textsc{Jambu} \citelanguageresource{aroraJambuHistoricalLinguistic2023} \\
        \multirow[t]{3}{7em}{Indo-European} & Indo-European Cognate Relationships (IE-CoR) Database \citelanguageresource{heggartyLanguageTreesSampled2023}, \\
        & Germanic \citelanguageresource{luoAutomaticMethodsSound2021}, Indo-European Lexicon (IELEX) \citelanguageresource{lrcIELEX2024}, \\ & \textsc{Jambu} \citelanguageresource{aroraJambuHistoricalLinguistic2023}, \textbf{PILA} [Italic] (Ours), Slavic \citelanguageresource{cathcartSearchIsoglossesContinuous2020} \\
        Mongolic-Khitan & \textsc{Jambu} [Mongolic] \citelanguageresource{aroraJambuHistoricalLinguistic2023} \\
        \multirow[t]{2}{7em}{Sino-Tibetan} & Bai$^\dag$ \citelanguageresource{fengwangLanguageContactLanguage2004}, Burmish$^\dag$ \citelanguageresource{gongMaterialsEtymologicalDictionary2020}, Karen$^\dag$ \citelanguageresource{luangthongkumViewProtoKarenPhonology2019}, \\
        & Lalo$^\dag$ \citelanguageresource{yangLaloRegionalVarieties2023}, Tujia$^\dag$ \citelanguageresource{zhouProtoBizicStudyTujia2020} \\
        Uto-Aztecan & Aztecan$^\dag$ \citelanguageresource{davletshinProtoUtoAztecansTheirWay2012}, Corachol and N\stress{a}huatl$^\dag$ \citelanguageresource{pharaohansenFamiliaVecinosInvestigando2020} \\
        Turkic & \textsc{Jambu} \citelanguageresource{aroraJambuHistoricalLinguistic2023} \\
        \bottomrule
    \end{tabular}
    \caption{A depiction of proto-form dataset coverage across language families. If datasets are named, their names are presented followed by the language subfamilies which they concern, if applicable, in square brackets. Otherwise, the name(s) of contained language subfamilies are used to represent the dataset. Items are marked with a $\dag$ if they are, at the time of writing, made available through Lexibank.}
    \label{tab:language-families}
\end{table*}

\begin{table*}[t]
    \footnotesize
    \centering
    \begin{tabular}{lllr}
        \toprule
        Dataset & Ancestor & Descendant & Pairs \\
        \midrule
        IELEX \citepaliaslanguageresource{lrcIELEX2024} & Proto-Indo-European (indo1319) & English (stan1293) & 14697 \\
        \citetlanguageresource{luoAutomaticMethodsSound2021} & Proto-Germanic (germ1287) & Old English (olde1238) & 4599 \\
        \textsc{Jambu} \citelanguageresource{aroraJambuHistoricalLinguistic2023} & Proto-Dravidian (drav1251) & Tamil (tami1289) & 4276 \\ 
        \textbf{PILA (Ours)} & Proto-Italic (ital1284) & Latin (lati1261) & 2916 \\
        \citetlanguageresource{cathcartSearchIsoglossesContinuous2020} & Proto-Slavic (slav1255) & Russian (russ1263) & 1572 \\ 
        MCD \citelanguageresource{benderProtoMicronesianReconstructions2003a,benderProtoMicronesianReconstructions2003b} & Proto-Chuukic (truk1243) & Chuukese (chuu1238) & 1460 \\
        \citetlanguageresource{yangLaloRegionalVarieties2023} & Proto-Lalo & Shuizhuping (west1506) & 954 \\
        \citetlanguageresource{fengwangLanguageContactLanguage2004} & Proto-Bai & Enqi (enqi1234) & 455 \\
        \citetlanguageresource{luangthongkumViewProtoKarenPhonology2019} & Proto-Karen (kare1337) & Northern Sgaw & 366 \\
        \citetlanguageresource{zhouProtoBizicStudyTujia2020} & Proto-Bizic & Cebu (sout2739) & 324 \\
        \citetlanguageresource{gongMaterialsEtymologicalDictionary2020} & Proto-Burmish (burm1266) & Maru (maru1249) & 264 \\
        \citetlanguageresource{decarvalhoComparativeReconstructionProtoPurus2021} & Proto-Purus (puru1265) & Yine (yine1238) & 201 \\
        IECoR \citelanguageresource{heggartyLanguageTreesSampled2023} & Proto-Indo-European (indo1319) & Old Polish (poli1260) & 146 \\
        \citetlanguageresource{chaconRevisedProposalProtoTukanoan2014} & Proto-Tucanoan (tuca1253) & Tukano (tuca1252) & 131 \\
        \citetlanguageresource{davletshinProtoUtoAztecansTheirWay2012} & Proto-Aztecan (azte1234) & Classical Nahuatl (clas1250) & 84 \\
        \citetlanguageresource{pharaohansenFamiliaVecinosInvestigando2020} & Proto-Nahua (azte1234) & Cora (cora1259) & 55 \\
        \bottomrule
    \end{tabular}
    \caption{A comparison of datasets containing proto-languages. Each language pairing presented is the pairing with the maximal number of etymon--reflex pairs for the given dataset. Each language is joined by its provided (or otherwise determinable) Glottocode. Reconstructions are included whether they are \textit{full} or \textit{partial} (as defined in \cref{sec:pila-trimming}), as it is nontrivial to distinguish between these reconstruction types.}
    \label{tab:proto-dataset-comparison}
\end{table*}

Recent interest in computational historical linguistics has spurred an uptick in the number of available datasets---including those with proto-languages. Work from Lexibank \citelanguageresource{listLexibankPublicRepository2022}, a large collection of historical linguistics datasets converted into the Cross-Linguistic Data Format (CLDF) \cite{forkelCrossLinguisticDataFormats2018}, has been central to the increase in accessible datasets across language families.

Nevertheless, the number of datasets with proto-languages remains small, and coverage of language families could be greatly improved. Lexibank contains languages from four of Glottolog's 245 language families \citelanguageresource{hammarstromGlottologGlottologGlottolog2023}. With the inclusion of all non-Lexibank datasets that we know of, a total of eight language families have received attention. We list all datasets containing proto-languages with cognate relationships in \cref{tab:language-families}. As this table illustrates, our work fills a gap, being the first dataset to provide explicit coverage for Italic. \cref{tab:proto-dataset-comparison} further shows that PILA is one of the largest datasets of its kind.\footnote{Because it does not differentiate attested Sanskrit and unattested proto-forms, an Old Indo-Aryan dataset \citelanguageresource{cathcartDisentanglingDialectsNeural2020} is left out from \cref{tab:language-families,tab:proto-dataset-comparison}.}

Outside of Lexibank, there are other datasets that document etymon--reflex relations between attested languages. For example, the WikiHan dataset compares Middle Chinese to eight Chinese subgroups \citelanguageresource{changWikiHanNewComparative2022}. Another group of related datasets compare Latin with five descendant Romance languages \citelanguageresource{ciobanuBuildingDatasetMultilingual2014,meloniAntiquoNeuralProtolanguage2021}. One of two Coglust (Cognate Clustering) datasets similarly deals with Latin and Romance languages, adding Catalan to their set; the other contains Turkic and six of its descendants \citelanguageresource{wuCreatingLargescaleMultilingual2018}. 

Yet other datasets have less of a focus on modeling the relationship between specific ancestor and descendant languages and instead identify cognate sets over a wide range of languages.
Namely, the Indo-European Lexicon (IELEX) \citelanguageresource{lrcIELEX2024}, 
the Indo-European Cognate Relationships (IE-CoR) database \citelanguageresource{heggartyLanguageTreesSampled2023}, CogNet \citelanguageresource{batsurenCogNetLargescaleCognate2019, batsurenLargeEvolvingCognate2022}, the \textsc{Jambu} database for South Asian languages \citelanguageresource{aroraJambuHistoricalLinguistic2023}, and a pair of cognate datasets descending from Proto-Germanic and Latin from \citeauthor{luoAutomaticMethodsSound2021} all contain over 100 languages.

\section{PILA: Proto-Italic to Latin Dataset}
\label{sec:pila}

\subsection{Overview}
\label{sec:pila-overview}

PILA is intended to document etymon--reflex relationships between Proto-Italic and Latin. However, languages change over time even as they bear the same name. Therefore, to assure consistency in etymon--reflex relationships, PILA specifically captures Proto-Italic and Latin at stages of development we call \enquote{Pre-Latin Proto-Italic} and \enquote{Ciceronian Latin} (see \cref{fig:partial-family-tree}). 

We selected the former to focus PILA on phonetic changes, as this stage of Proto-Italic has already undergone many non-phonological changes which would obscure sound change laws. As for Ciceronian Latin, we chose it not only because it serves as a standard version of Latin (promoted by the famed orator, Cicero) but also because its distance from Pre-Latin Proto-Italic allows for many meaningful phonetic changes to have occurred.

To store our dataset, we use the Cross-Linguistic Data Format (CLDF) \cite{forkelCrossLinguisticDataFormats2018}. This format consists of CSV, JSON, and Bib\TeX{} files that conform to a specification built on the CSV on the Web (CSVW) standard. By providing explicit and uniform structural requirements, CLDF enforces a principled style of data table design and allows adherent datasets to readily work with libraries supporting the data format (\eg, the historical linguistics library LingPy \cite{listLingPyPythonLibrary2021}).

\begin{table}[t]
    \small
    \centering
    \begin{tabular}{@{}lrrr@{}}
        \toprule
        & Latin & Proto-Italic & All \\
        \midrule
        \# Forms & 2860 & 2916 & 5776 \\
        \# Phones & 15974 & 18779 & 34753 \\
        \# Phone Types & 33 & 41 & 48 \\  
        \midrule
        Avg. Length & $5.6\,\pm\,1.4$ & $6.4\,\pm\,1.8$ & $6.0\,\pm\,1.7$ \\
        \bottomrule
    \end{tabular}
    \caption{Collection of PILA dataset statistics relative to its component languages. The first section of the table consists of frequencies, whereas the second contains averages with standard deviations.}
    \label{tab:dataset-statistics}
\end{table}

Our dataset builds on CLDF's \code{Wordlist} module. Disregarding the dataset's JSON table metadata file and its Bib\TeX{} file to store cited sources, seven tables (CSV files) constitute the dataset:

\begin{enumerate}
    \item \code{languages.csv}: a collection of languages contained in the dataset and their attributes.
    \item \code{forms.csv}: a collection of phonetic sequences and tokenized phones. It contains identifiers to link forms with all other tables in PILA. Statistics for these forms are collected in \cref{tab:dataset-statistics}.
    \item \code{cognates.csv}: a collection of numeric identifiers which link each form to its cognate set.
    \item \code{lemmata.csv}: a collection of lemmata to which our forms are morphologically related. Currently, only Latin forms have lemmata.
    \item \code{glosses.csv}: a collection of notes and tabulated irregular phenomena for our forms. For our irregularity categories, see \cref{tab:irregularity-table}.
    \item \code{tags.csv}: a collection of groups of morphological tags which correspond to Latin forms in our dataset. See \cref{sec:augmentation-inflections} for details.
    \item \code{overlaps.csv}: a collection of identifiers linking forms in PILA and in others' datasets together. See \cref{sec:dataset-compatibility} for more details.
\end{enumerate}

\subsection{Development}
\label{sec:pila-development}

In this section, we describe our dataset development procedure. We worked closely with an expert in historical linguistics for Proto-Italic and Latin (the last author) and a graduate student with a decade of Latin experience (the first author) to scrape, trim, normalize, augment, and annotate the data. We describe each step below. Although this procedure is framed in the context of Latin and Proto-Italic, many of its steps (\eg, our scraping procedure) could be performed with minor contextual changes for any language pair. 

\subsubsection{Scraping}
\label{sec:pila-scraping}

We initially extracted data from Wiktionary. Because of public availability and decent etymological curation, Wiktionary was deemed to be a suitable starting point for PILA. We scraped all pages tagged as \enquote{Latin terms derived from Proto-Italic} for their Latin headwords and Proto-Italic forms (tagged with the \code{lang} attribute \enquote{itc-pro}) \citelanguageresource{wiktionarycontributorsCategoryLatinTerms2017}.\footnote{The scraping was done on February \nth{15}, 2022.} For each \enquote{etymology} heading in the Latin section of a headword's page, we extracted an etymon--reflex pair. (Multiple etymologies are possible if multiple Latin words happen to have the same spelling.) We automatically cleaned the natural-language formatting of etymologies. We filtered out affix headwords (as they do not occur in natural speech), mislabeled headwords, and reconstructed Latin headwords from our data.

\subsubsection{Trimming}
\label{sec:pila-trimming}

We manually classified etymon-reflex pairs as \textit{partial} or \textit{full} reconstructions. Partially-reconstructed pairs have an etymon either with mismatching parts of speech or with mismatching or missing morphemes. For example, Wiktionary reported (at the time of data collection) an etymology for the Latin verb \latin{audeo\tricolon} \gloss{I dare} as the Proto-Italic adjective \pital{*awidos}. Meanwhile, a fully-reconstructed pair has matching parts of speech and morphemes; in PILA, we have such pairs in \pital{*awide\tricolon o\tricolon} and \latin{audeo\tricolon} and also in \pital{*awi\dh os} and \latin{avidus} \gloss{desirous}.
Having fully-reconstructed forms is desirable because these forms fully represent phonological developments with respect to both stems and affixes, whereas partially-reconstructed forms do not. Therefore, we dropped the partially-reconstructed forms, producing a set of 1,205 fully-reconstructed pairs.

Once these pairs were gathered, our experts examined them for quality. All Proto-Italic forms were checked against standard etymological dictionaries and grammars \cite{sihlerNewComparativeGrammar1995,meiserHistorischeLautUnd2010,leumannLateinscheLautUnd1977,waldeLateinischesEtymologischesWorterbuch1938,ernoutDictionnaireEtymologiqueLangue2001}, such as the Etymological Dictionary of Latin (henceforth EDL) \cite{devaanEtymologicalDictionaryLatin2008}. This resulted in the deletion of various forms. Such forms largely fell into one of two categories. 

First, a form could have either fallen out of use before our Ciceronian Latin period or could have become attested only after that period. Examples of these include \latin{aevita\tricolon s} \gloss{age, lifetime}, which was superseded by \latin{aeta\tricolon s}, for early attestation and \latin{genimen} \gloss{progeny} for late attestation. Second, a form could have been a proper name. Proper names have a tendency to contain non-Latin or non-Indo-European phonetic sequences, diverging from PILA's focus on Italic phonetics. An example is the Latin \latin{ti\tricolon bur}, the name of a town.

We also made minor alterations to some forms to reflect known Latin developments. These tend to have case-by-case explanations, which we store among our notes in the \code{glosses.csv} table.

\subsubsection{Normalization}
\label{sec:pila-normalization}

We normalized the Latin and Proto-Italic data to consistent phonetic representations. This step was necessary because Wiktionary contributors followed different standards in different entries. 

Proto-Italic forms were normalized to EDL's conventions, as this work is a current exemplar for Latin historical linguistics \cite{devaanEtymologicalDictionaryLatin2008}. However, a few alterations were made to improve the phonetic representations. First, because voicing assimilation appears to be a consistent feature of Pre-Latin Italic, \pital{*s} before a voiced consonant or between vowels was written as \pital{*z}. For example, the etymon for the adjective \latin{numerus} \gloss{number} is \pital{*nomezos} instead of \pital{*nomesos}. Second, fricatives that resulted from the loss of PIE's voiced aspirates were reconstructed as voiceless in initial position and as voiced between vowels or before a voiced consonant. The word \latin{faber} \gloss{craftsman} exhibits this well: its etymon, \pital{*fa\textbeta{}er}, has an initial voiceless fricative (\pital{*f}) and a medial voiced fricative (\pital{*\textbeta{}}).

Meanwhile, for Latin, we largely preserved its orthography but again made alterations when the orthography hid the actual pronunciation. We changed \latin{n} or \latin{g} to \latin{\ng} where they were actually pronounced as \latin{\ng}. For instance, \latin{magnus} \gloss{great} was written as \latin{ma\ng nus} and \latin{frango\tricolon} \gloss{I shatter} as \latin{fra\ng go\tricolon}. Similarly, we duplicate \latin{i} when the orthography conceals that it is used both as a vowel and as a consonant, as in \latin{maiior} \gloss{greater} and \latin{cuiius} \gloss{whose}.

\subsubsection{Augmentation: Novel Forms}

\begin{table*}[t]
    \centering
    \small
    \begin{tabularx}{\linewidth}{@{}lX|ccc@{}}
        \toprule
        \textbf{Category} & \textbf{Definition} & \multicolumn{3}{c}{\textbf{Example}} \\
        & & Etymon & Expected & Actual \\
        \midrule
        Association & A form with a change attributable to its association with another word or semantic set. & \pital{*kleiments} & \latin{cli:me:ns} & \latin{cle:me:ns} \\
        Borrowing & A form with a sound change different from standard Latin and/or from another Italic dialect. & \pital{*g\textsuperscript{w}o:s} & \latin{vo:s} & \latin{bo:s} \\
        Morphology & A form where morphosemantics undid, blocked, or otherwise interfered with expected sound changes. & \pital{*mensen} & \latin{me:nsen} & \latin{me:nsis} \\
        Paradigm Leveling & A form which exhibits changes taken from another part of its own paradigm or lexical family. & \pital{*weznos} & \latin{ve:nus} & \latin{ve:ris} \\
        Phonology & A form with an uncommon or unexpected phonological phenomenon. & \pital{*dikitos} & \latin{dicitus} & \latin{digitus} \\
        \bottomrule
    \end{tabularx}
    \caption{PILA sound change irregularity categories, definitions, and examples.}
    \label{tab:irregularity-table}
\end{table*}

We added a variety of forms to the cleaned, fully-reconstructed set. In consultation with EDL \cite{devaanEtymologicalDictionaryLatin2008}, we completed some of the partial reconstructions and appended additional forms. Some appended forms were chosen by our experts to provide representation for missing morphological features. Specifically, the scraped pairs did not include any perfect-tense forms. Thus, we added some in both the active and passive voices.

The perfect active \enquote{third principal part} for standard Latin verbs largely falls into one of four main categories: reduplicating perfects (\eg, \latin{fefelli\tricolon} \gloss{I deceived} for \latin{fallo\tricolon} \gloss{I deceive}); root aorists (\eg, \latin{ru\tricolon pi\tricolon} \gloss{I broke} for \latin{rumpo\tricolon} \gloss{I break}); \latin{s} aorists (\eg, \latin{scri\tricolon psi\tricolon} \gloss{I wrote} for \latin{scri\tricolon bo\tricolon} \gloss{I write}); and \latin{vi\tricolon}/\latin{ui\tricolon} perfects (\eg, \latin{ama\tricolon vi\tricolon} \gloss{I loved} for \latin{amo\tricolon} \gloss{I love} or \latin{monui\tricolon} \gloss{I warned} for \latin{moneo\tricolon} \gloss{I warn}). Some of these forms pose problems.

Regarding reduplicating perfects, the inherited reduplicating vowel is \latin{e} (\eg, \latin{memordi\tricolon} \gloss{I bit} for \latin{mordeo\tricolon} \gloss{I bite}). Apparently, when a verbal root itself contained an \pital{*e}, the generative principle for the reduplicated syllable was reinterpreted to incorporate the vowel of the root instead of \pital{*e} with the initial consonant. That became the standard pattern in reduplicated perfects (\eg, \latin{cucurri\tricolon} \gloss{I ran}---and not \latin{cecurri:}---for \latin{curro\tricolon} \gloss{I run}), and we reconstructed reduplicating perfects accordingly. %; \latin{momordi\tricolon} replacing \latin{memordi\tricolon}---there is no known sound law that would make \latin{memor-} into \latin{momor-}). 

Meanwhile, concerning \latin{vi:}/\latin{ui:} perfects, their origins are not clear. Latin is the only Italic language that uses that suffix for the perfect. A form like \pital{*monawai} producing \latin{monui\tricolon} may therefore not be a truly Proto-Italic feature but a late, \enquote{Pre-Latin} one. However, given the importance of these perfects in the Latin system and the Pre-Latin character of our Proto-Italic, we decided to include them.

Turning to perfect passive forms, we added the Latin paradigm-defining \pital{*to} participle. In some cases this participle's form is due to paradigm leveling (see \cref{tab:irregularity-table}) or other types of analogical change (\eg, perhaps \latin{pulsus} \gloss{having been struck} after the perfect active \latin{pulsi\tricolon} \gloss{I struck}). Such forms were excluded, and older forms, where known or securely reconstructable, were used (\eg, \latin{pultus}, which must have existed, as it is the base of the frequentative \latin{pulta\tricolon re} \gloss{to knock}).

Through this process, we created a set of 1,515 Latin forms and 1,548 Proto-Italic forms. 

\subsubsection{Augmentation: Inflections}
\label{sec:augmentation-inflections}

We elected to further augment this data with inflected forms. This seemed crucial for our target languages, as they are rife with inflections.
Moreover, sound changes may occur differently in the citation form and inflected forms of a word. For instance, the Latin \latin{cru\tricolon s} \gloss{leg} has a genitive form \latin{cru\tricolon ris} \gloss{of the leg}. The final \latin{s} in \latin{cru\tricolon s} and the medial \latin{r} in \latin{cru\tricolon ris} arise from the same original \pital{*s}; however, because the genitive suffix, \latin{is} (Proto-Italic \pital{*es}), begins with a vowel, the stem's \pital{*s} found itself between two vowels, causing it to rhotacize. Many other Latin forms display similar behaviors (\eg, \latin{flo\tricolon s} \gloss{flower} and \latin{flo\tricolon ris} \gloss{of the flower}). 

Because nouns, adjectives, participles, and verbs are the primary subjects of inflection in Latin, we added inflections for words in those categories.

\paragraph{Nominal Forms}

To most nouns, adjectives, and participles, we added genitive singular forms. In all declensions, these endings pose certain difficulties.

In the first or \latin{a\tricolon} declension, Latin inherited the PIE ending \pital{*a\tricolon s}, preserved in archaic phrases and older texts. But the analogical ending \pital{*a\tricolon i\tricolon}, with \pital{*i\tricolon} imported from the second declension, became the standard ending. We used that ending, as it is the result of non-phonological changes that lead into Pre-Latin Proto-Italic. Meanwhile, in the second or \latin{o} declension, Latin preserves the inherited ending \pital{*osyo} quite rarely. In attested Latin, the ending is \latin{i\tricolon}; thus, we used that ending in Proto-Italic as well.

Lastly, the Latin third declension represents a fusion of PIE i-stems and consonant stems. These i-stems affixed consonant stem endings to an ablauting medial \pital{*ej}. 
The earlier Latin i-stem genitive ending was \latin{i\tricolon s} from \pital{*ejes}. But that affix was replaced by the \latin{is} of the consonant stems. 
That affix, in turn, represents the generalization of the outcome of inherited \pital{*es} over the outcome of inherited \pital{*os} (attested, very rarely, as \latin{us}). 
All Proto-Italic i-stems and consonant stems were thus reconstructed with genitive \pital{*es}.

\paragraph{Verbal Forms}

The \nth{3}-person singular present of all indicative verbs was added. It is, on some accounts, the least marked member of a paradigm. A problem is posed by the so-called \textit{hic et nunc} (\enquote{here and now}) marker \pital{*i}, which tagged present tense forms in PIE but disappeared in Proto-Italic. The contrast in Old Latin third person endings \pital{*ti} and/or \pital{*t} producing Latin \latin{t} and/or \latin{d} suggests that the origin of \latin{t} is (partly) phonological. Therefore, verbs were reconstructed with that ending.

\paragraph{Morphological Tags}

To provide a way to analyze PILA with respect to its inflections, we incorporate a set of morphological tags into our dataset through the \code{tags.csv} file. This file lists common morphological properties for all our Latin forms. We adapted the tagset used by the Perseus Project's morphological analyzer, Morpheus \cite{craneGeneratingParsingClassical1991}, for our purposes. Namely, we considered the part-of-speech, person, number, tense, mood, voice, gender, case, and degree for each form. Furthermore, we added a tag for each form's \textit{inflection class}---that is, the paradigm of conjugation or declension to which a word belongs (if any).

Although we largely adhered to Morpheus' tagset, we made a few adjustments to suit the ambiguity that comes with examining forms devoid of sentential (and, thus, syntactic) context. For instance, although many Latin adjectives change their inflection class to conform to the gender of the noun that they modify, some---such as the word \latin{fe\tricolon li\tricolon x} \gloss{fortunate}---do not change at all. As a result, we include all possible combinations of genders as potential tags. We detail other aspects of our tagging scheme in our dataset repository.

\subsubsection{Annotation}

To account for the influence of irregular (\ie, rare or non-phonological) changes in phonological studies, we annotated PILA's etymon--reflex pairs with tags to flag the presence of such changes.
We present the categories for these tags in \cref{tab:irregularity-table}. In our dataset, a gloss accompanies each tag to justify its attachment to a given form. One application of these tags could be to filter PILA to a set of forms which have undergone a specific kind of change to study the effects of that change.

\section{Applications}
\label{sec:applications}

So far, we have detailed the PILA dataset. In this section, we exhibit PILA's applicability to standard computational historical linguistics studies (\cref{sec:sample-tasks}) and PILA's capacity to enhance existing datasets through overlapping forms (\cref{sec:dataset-compatibility}).\footnote{The code for these applications, as well as for various dataset creation and analysis utilities, is available at this location: \url{https://github.com/Mythologos/PILA-Code}.}

\subsection{Sample Tasks}
\label{sec:sample-tasks}

In this section, we perform two traditional computational historical linguistics tasks. Suppose that we have a pair of languages, an ancestor $E$ and a descendant $R$, both considered as sets of forms. Then we can define a set of etymon--reflex pairs $C \subseteq E \times R$. We note that both $e \in E$ and $r \in R$ could be used in more than one pair.

For this set $C$, we can define two tasks. First, we define \textit{reflex prediction} as the task of predicting $r$, given $e$. Second, we define \textit{etymon reconstruction} as the task of predicting $e$, given $r$. We perform both of these tasks with PILA below.

\subsubsection{Procedure}
\label{sec:sample-task-procedure}

To perform the reflex prediction and etymon reconstruction tasks, we grouped our data by lemma (citation form) to ensure that inflected forms of the same lemma stay together, and we randomly split the lemmata into training (80\%), validation (10\%), and test (10\%) splits. This resulted in sets of 2,331, 298, and 287 etymon--reflex pairs, respectively.

We train a Transformer encoder--decoder model \cite{vaswaniAttentionAllYou2017a}, and we match most of the hyperparameter settings used in that work. We use PyTorch's Transformer layer implementation \citep{paszkePyTorchImperativeStyle2019}.
Unlike \citet{vaswaniAttentionAllYou2017a}, we use pre-norm instead of post-norm \citep{wangLearningDeepTransformer2019,nguyenTransformersTearsImproving2019} and apply layer normalization to the last layer's output.
We use 6 layers in both the encoder and decoder and 8 attention heads per layer.
We initialize the output layer with Xavier uniform initialization \cite{glorotUnderstandingDifficultyTraining2010}. For layer norm, we initialize all weights to 1 and all biases to 0. We initialize all other parameters by sampling uniformly from $[-0.01, 0.01]$.

We optimize parameters with Adam \citep{kingmaAdamMethodStochastic2015a}. We clip gradients with a threshold of~5 using $L^2$ norm rescaling. We train the model by minimizing the decoder's cross-entropy (summed over all timesteps) on the training set. We use label smoothing with a weight of 0.1. We take a checkpoint every 2,000 examples to evaluate the decoder's per-token cross-entropy on the validation set. After two checkpoints with no improvement, we multiply the learning rate by 0.5; after two more such checkpoints, we stop training early. We train for up to 100 epochs. We use the checkpoint with the best validation cross-entropy.

For each epoch, we randomly shuffle examples and group examples of similar lengths into the same minibatch. We limit the number of tokens in the source or target side of a batch to $B$ tokens, including padding, \bos{}, and \eos{} symbols. We use beam search decoding with a beam size of 4. We apply length normalization to hypothesis probabilities before selecting the top $k$ hypotheses for the next beam. We do this by dividing the log-probability of each hypothesis by the number of tokens generated by that hypothesis so far (including \eos{}).

We perform a random hyperparameter search \cite{bergstraRandomSearchHyperparameter2012} over 10 runs. For each run, we randomly sample four hyperparameters: the initial learning rate, the batch size $B$, the model size $s$, and the dropout rate. See \cref{tab:hyperparameter-spaces} for our chosen distributions. With the model size, we set $\dmodel$ to $8 \cdot s$ and the size of the feedforward hidden layers to $4 \cdot \dmodel$. We apply dropout as PyTorch does, which follows \citet{vaswaniAttentionAllYou2017a} and also applies it to feedforward sublayers' hidden units and attention probabilities.

To evaluate our models, we use multi-reference word error rate (WER) and phoneme error rate (PER). \enquote{Multi-reference} means that when there are multiple references (multiple etyma for one reflex), we take the minimum error across all references. Note that for WER, the number of errors is either 0 or 1, so WER is one minus the exact match rate. For PER, we use micro-averaging: we sum the total number of edits and total number of reference symbols over the whole test set, reporting their ratio.

\begin{table}[t]
    \centering
    \small
    \begin{tabular}{@{}lll@{}}
        \toprule
        Hyperparameter & Distribution & Range \\
        \midrule
        Batch Size & Uniform & $[32, 256]$ \\
        Dropout Rate & Uniform & $[0, 0.2]$ \\
        Learning Rate & Log-Uniform & $[0.0001, 0.01]$ \\
        Model Size & Uniform & $[4, 64]$ \\ 
        \bottomrule
    \end{tabular}
    \caption{Collection of hyperparameters and search spaces used in our sample tasks.}
    \label{tab:hyperparameter-spaces}
\end{table}

\subsubsection{Results}
\label{sec:sample-task-results}

\begin{table*}[t]
    \centering
\begin{tabular}{l|cc|cc}
\toprule
& \multicolumn{2}{c|}{Proto-Italic $\rightarrow$ Latin} & \multicolumn{2}{c}{Proto-Italic $\leftarrow$ Latin} \\
Model & PER ($\downarrow$) & WER ($\downarrow$) & PER ($\downarrow$) & WER ($\downarrow$) \\
\midrule
Copying & 0.53 & 0.98 & 0.46 & 0.97 \\
Transformer & 0.18 & 0.52 & 0.24 & 0.73 \\
\bottomrule
\end{tabular}
    \caption{Results for phone prediction tasks on PILA's test set. The \enquote{Copying} baseline copies the input to the output. All \enquote{Transformer} results are derived from the best model from our hyperparameter search.}
    \label{tab:prediction-results}
\end{table*}

\begin{table*}[t]
    \centering
\begin{tabular}{lcccc}
\toprule
Task & $\dmodel$ & Dropout Rate & Learning Rate & Batch Size $B$ \\
\midrule
Proto-Italic $\rightarrow$ Latin & 112 & 0.1665 & 0.00021969 & 138 \\
Proto-Italic $\leftarrow$ Latin & 496 & 0.0875 & 0.00010027 & 81 \\
\bottomrule
\end{tabular}
    \caption{Randomly-searched hyperparameters of the best Transformer models from \cref{tab:prediction-results}.}
    \label{tab:prediction-hyperparameters}
\end{table*}

In \cref{tab:prediction-results}, we report results on the test set from the hyperparameter search's best-performing model on the validation set. Corresponding hyperparameters for the best models are shown in \cref{tab:prediction-hyperparameters}.

We compare the Transformer model discussed in the previous section to a \enquote{Copying} baseline. In this baseline, the input is simply copied to the output. This baseline is somewhat reasonable because, although Proto-Italic phones undergo many sound changes in becoming Latin, some remain recognizably unchanged. 
Thus, by comparing a model to the \enquote{Copying} baseline, we examine whether it learns any nontrivial sound change rules. As \cref{tab:prediction-results} shows, the Transformer baseline vastly outperforms the \enquote{Copying} baseline in both directions, indicating that this is the case---and that, in fact, PILA provides a learnable signal.

\subsection{Dataset Compatibility}
\label{sec:dataset-compatibility}

In this section, we show to what extent PILA's entries can be linked to those of other datasets, allowing for models to be built with longer chains of sound changes and additional linguistic metadata.

To measure PILA's capacity to link to other datasets, we tally the overlap between PILA's and other datasets' Latin forms. Because datasets organize data differently, it is nontrivial to extract overlap counts. Moreover, because datasets vary in their attention to phonetic features such as vowel length, the legitimacy of matches can be murky (\eg, without vowel length, PILA's adjectives \latin{le\tricolon vis} \gloss{smooth} and \latin{levis} \gloss{light} would be indistinguishable).

To account for this, we define two categories of overlap. We say that an overlap is \textit{direct} if phonological or morphological information is not lost in performing the match. Conversely, an overlap is \textit{indirect} if some such information is lost. For instance, a form may need to be inflected differently, resulting only in a partial compatibility (or perhaps a false positive match due to homography) between the information stored in each dataset.

\subsubsection{Procedure}

\begin{algorithm}[t]
    \caption{Dataset Compatibility Algorithm}
    \begin{algorithmic}[1]
        \Procedure{scoreComp}{first, second}
            \State $M, N \gets |\text{first}|, |\text{second}|$
            \State $\text{maxSize} \gets \max(M, N)$
            \State $\text{scores} \gets \text{zeros}(\text{maxSize}, \text{maxSize})$
            \For {$i \in 1\ldots M$}
                \For {$j \in 1\ldots N$}
                    \If {$\text{first}[i]\bigcap\text{second}[j] \neq \emptyset$}
                        \State $scores[i][j] \gets 1$
                    \Else
                        \State $scores[i][j] \gets 0$
                    \EndIf
                \EndFor
            \EndFor
            \State maxEntries $\gets$ LSA(scores)
            \State score $\gets$ 0
            \For{$(i, j) \in \text{maxEntries}$}
                \State $\text{score} \gets \text{score} + \text{scores}[i][j]$
            \EndFor
            \State \textbf{return} score
        \EndProcedure
    \end{algorithmic}
    \label{alg:dataset-compatibility-algorithm}
\end{algorithm}

\begin{table*}
    \centering
    \small
    \begin{tabular}{l|r|rrrr}
        \toprule
         Dataset & Latin Forms & \multicolumn{1}{c}{Direct} & \multicolumn{1}{c}{Indirect} & \multicolumn{1}{c}{Total} \\
         \midrule \citetlanguageresource{ciobanuBuildingDatasetMultilingual2014} & 3218 & 147 \halfquad(4.6\%) & 31 \halfquad(0.1\%) & 178 \halfquad(5.5\%) \\ \citetlanguageresource{meloniAntiquoNeuralProtolanguage2021} -- Additions & 5419 & 68 \halfquad(1.3\%) & 580 (10.7\%) & 648 (12.0\%) \\ \citetlanguageresource{meloniAntiquoNeuralProtolanguage2021} -- Full & 8799 & 135 \halfquad(1.5\%) & 847 \halfquad(9.6\%) & 982 (11.2\%) \\
         \midrule
         Coglust \citelanguageresource{wuCreatingLargescaleMultilingual2018} & 27645 & 760 \halfquad(2.8\%) & 521 \halfquad(1.9\%) & 1281 \halfquad(4.6\%) \\
         CogNet \citelanguageresource{batsurenCogNetLargescaleCognate2019,batsurenLargeEvolvingCognate2022} & 6960 & 354 \halfquad(5.1\%) & 458 \halfquad(6.6\%) & 812 (11.7\%) \\
         IE-CoR \citelanguageresource{heggartyIndoEuropeanCognateRelationships2022} & 266 & 97 (36.5\%) & 44 (16.5\%) & 141 (53.0\%) \\
         IELEX \citelanguageresource{lrcIELEX2024} & 10110 & 558 \halfquad(5.5\%) & 621 \halfquad(6.1\%) & 1179 (11.7\%) \\
         \textsc{Jambu} \citelanguageresource{aroraJambuHistoricalLinguistic2023} & 4 & 2 (50.0\%) & 0 \halfquad(0.0\%) & 2 (50.0\%) \\
         \citetlanguageresource{luoAutomaticMethodsSound2021} -- Romance & 10866 & 489 \halfquad(4.5\%) & 676 \halfquad(6.2\%) & 1165 (10.7\%) \\
         \bottomrule
    \end{tabular}
    \caption{Dataset compatibility study results. Columns measure degrees of overlap relative to PILA. Integers are counts; percentages are relative to the number of data points in that row's dataset. The top three datasets are from related works and are comprised of similar data.}
    \label{tab:dataset-compatibility-results}
\end{table*}

\Cref{alg:dataset-compatibility-algorithm} sketches our overlap computation procedure. This procedure centers around the Hungarian or Kuhn--Munkres algorithm \cite{kuhnHungarianMethodAssignment1955, munkresAlgorithmsAssignmentTransportation1957}, which solves the \textit{linear sum assignment} problem. In our pseudocode, this algorithm is named \code{LSA}, and in practice we use SciPy's implementation \cite{virtanenSciPyFundamentalAlgorithms2020}. Given a weighted bipartite graph, the Hungarian algorithm
seeks the one-to-one matching that maximizes the sum of those edges' weights. 

The function $\textproc{scoreComp}$ takes as arguments two lists of nodes (one for each dataset), and each node contains one or more forms.
We create a bipartite graph whose nodes are the aforementioned nodes, and if node $u$ and node $v$ have some form in common, there is an edge between $u$ and $v$ with weight 1.
Applying a linear-sum assignment algorithm results in a list of edges. The number of edges is called the \emph{overlap}.

We apply this algorithm twice. First, we take both normalized datasets, place each form in its own node, and apply \cref{alg:dataset-compatibility-algorithm} to obtain the \emph{direct overlap}. Then, we remove all entries from both datasets that participated in the first matching. We take each dataset's remaining nodes, remove long marks (\texttt{\tricolon}) from each form, and add that form back to each node. For PILA's nodes exclusively, we use Collatinus \cite{ouvrardCollatinusOutilPolymorphe2014, verkerkCollatinusConvergenceLexica2020} as integrated into the Classical Language Toolkit \cite{johnsonClassicalLanguageToolkit2021}, to generate and include other inflections of each node's form.\footnote{%We remove long marks because Collatinus does not work with them. 
We also correct an error where Collatinus only generates inflected endings without any stem.}
We apply \cref{alg:dataset-compatibility-algorithm} again to get the \emph{indirect overlap}. Finally, we add the direct and indirect overlap to obtain the total overlap.

We examine nine other datasets which contain Latin forms. We also searched for fully-reconstructed Proto-Italic forms, but we found no dataset with any such forms.

\subsubsection{Results}

We present the results of our study for direct and indirect overlap in \cref{tab:dataset-compatibility-results}. In general, we find that PILA can match over 100 forms among all datasets that we examine (except for \textsc{Jambu}, as it contains less than 100 Latin forms), indicating a nontrivial overlap between PILA and other datasets. 

While direct matches generally account for most overlapping forms, indirect matches are prominent for \citeauthor{meloniAntiquoNeuralProtolanguage2021}'s datasets, for CogNet \citelanguageresource{batsurenCogNetLargescaleCognate2019,batsurenLargeEvolvingCognate2022}, for IELEX \citepaliaslanguageresource{lrcIELEX2024}, and for \citeauthor{luoAutomaticMethodsSound2021}'s Romance cognate dataset . For \citeauthor{meloniAntiquoNeuralProtolanguage2021}'s datasets, this is because they use the accusative singular inflection for nouns, adjectives, and participles and the present infinitive for verbs as their headwords, as these inflections more strongly link the Latin forms to their Romance language counterparts. Meanwhile, for the other datasets, many matches stem from either the removal of long marks, as their use is inconsistent, or the addition of morphologically-related forms.

In light of this study's results, we created an additional \code{overlaps.csv} table for PILA that facilitates sharing data between datasets. This table relates the indices of matched forms and designates the type of matching attained between them.

\section{Conclusion}
\label{sec:conclusion}

This paper introduced PILA, a historical-phonological dataset of etymon--reflex pairs in Proto-Italic and Latin. It described PILA's development process and organization. It provided baseline results for PILA on two historical linguistics tasks and showed PILA's capacity to enhance other datasets in a compatibility study.

Future work could expand the scope of PILA to encourage deeper historical linguistics studies. For instance, it could broaden its coverage of languages in the Italic region. Although they have scant extant data, languages like Umbrian \citelanguageresource{dehouckIKUVINATreebank2022} as well as Cisalpine Celtic, Faliscan, Oscan, and Venetic \cite{muranoDescribingInscriptionsAncient2023} have received some attention in computational literature.

In another direction, PILA could incorporate sets of phonological rules to support the task of automatic sound law induction (ASLI) \citelanguageresource{luoAutomaticMethodsSound2021,changAutomatingSoundChange2023}. However, many issues arise when considering how to select and store such rules. For example, what formalism should be used to organize sound change rules? Although historical linguists have a standard notation for sound laws, certain features and complex conditions do not have agreed-upon, computationally-friendly notation \cite{luoAutomaticMethodsSound2021}. An examination of prior ASLI work and the sound law databases UNIDIA \citelanguageresource{hamedDatabaseDerivingDiachronic2009} and PBase \citelanguageresource{mielkeEmergenceDistinctiveFeatures2008} could serve as a starting point for this direction.

\section{Acknowledgements}
\label{sec:acknowledgements}
Regarding the datasets used in this work, we would like to thank Todd Krause for providing us with a version of the IELEX dataset \citelanguageresource{lrcIELEX2024}. We would also like to thank Alina Maria Ciobanu and Liviu P. Dinu for allowing us to use their dataset \citelanguageresource{ciobanuBuildingDatasetMultilingual2014}, as well as Shauli Ravfogel for providing us with their revisions to it \citelanguageresource{meloniAntiquoNeuralProtolanguage2021}.

For their helpful comments and discussions concerning this work, we would also like to thank Darcey Riley, Ken Sible, Aarohi Srivastava, Chihiro Taguchi, and Andy Yang.

\section{Bibliographical References}
\label{sec:bibliography}
\vspace*{-4ex}
\bibliographystyle{lrec-coling2024-natbib}
\bibliography{pila}

\section{Language Resource References}
\label{sec:language-resource-refs}
\vspace*{-4ex}
\bibliographystylelanguageresource{lrec-coling2024-natbib}
\bibliographylanguageresource{languageresource}

\end{document}